\documentclass[10pt,twocolumn,letterpaper]{article}

\usepackage{wacv}              

\usepackage{graphicx}
\usepackage{amsmath}
\usepackage{amssymb}
\usepackage{booktabs}
\usepackage{multirow}
\usepackage{bbm}
\usepackage[pagebackref,breaklinks,colorlinks]{hyperref}
\usepackage[accsupp]{axessibility}

\usepackage[capitalize]{cleveref}
\crefname{section}{Sec.}{Secs.}
\Crefname{section}{Section}{Sections}
\Crefname{table}{Table}{Tables}
\crefname{table}{Tab.}{Tabs.}

\def\wacvPaperID{*****} 
\def\confName{WACV}
\def\confYear{2024}

\begin{document}

\title{Discriminator-free Unsupervised Domain Adaptation for Multi-label Image Classification}

\author{Inder Pal Singh$^1$, Enjie Ghorbel$^1$$^2$, Anis Kacem$^1$, Arunkumar Rathinam$^1$, Djamila Aouada$^1$\\
$^1$Interdisciplinary Centre for Security, Reliability and Trust (SnT), University of Luxembourg, Luxembourg\\
$^2$High Institute of Multimedia Arts (ISAMM), University of Manouba, Tunisia\\
{\tt\small \{inder.singh, enjie.ghorbel, anis.kacem, arunkumar.rathinam, djamila.aouada\}@uni.lu}
}
\maketitle

\begin{abstract}
In this paper, a discriminator-free adversarial-based Unsupervised Domain Adaptation (UDA) for Multi-Label Image Classification (MLIC) referred to as DDA-MLIC is proposed. Recently, some attempts have been made for introducing adversarial-based UDA methods in the context of MLIC. 
However, these methods, which rely on an additional discriminator subnet present one major shortcoming. The learning of domain-invariant features may harm their task-specific discriminative power, since the classification and discrimination tasks are decoupled.
Herein, we propose to overcome this issue by introducing a novel adversarial critic that is directly deduced from the task-specific classifier. Specifically, a two-component Gaussian Mixture Model (GMM) is fitted on the source and target predictions in order to distinguish between two clusters. This allows extracting a Gaussian distribution for each component. The resulting Gaussian distributions are then used for formulating an adversarial loss based on a Fr\'echet distance. The proposed method is evaluated on several multi-label image datasets covering three different types of domain shift. The obtained results demonstrate that DDA-MLIC outperforms existing state-of-the-art methods in terms of precision while requiring a lower number of parameters. The code is publicly available at \url{github.com/cvi2snt/DDA-MLIC}.\footnote{This research was funded in whole, or in part, by the Luxembourg National Research Fund (FNR), grant references BRIDGES2020/IS/14755859/MEET-A/Aouada and BRIDGES2021/IS/16353350/FaKeDeTeR. For the purpose of open access, and in fulfillment of the obligations arising from the grant agreement, the author has applied a Creative Commons Attribution 4.0 International (CC BY 4.0) license to any Author Accepted Manuscript version arising from this submission.}
\end{abstract}
\section{Introduction}
\label{sec:intro}
Multi-Label Image Classification (MLIC) aims at predicting the presence/absence of a set of objects in a given image. It is widely studied in the Computer Vision community due to its numerous fields of applications such as object recognition~\cite{multi-object}, scene classification~\cite{deeply}, and attribute recognition~\cite{human-attr, deepfake}.
\begin{figure}[!t]
\centering
\begin{subfigure}[b]{0.5\textwidth}
\centering
    \includegraphics[width=0.7\linewidth]{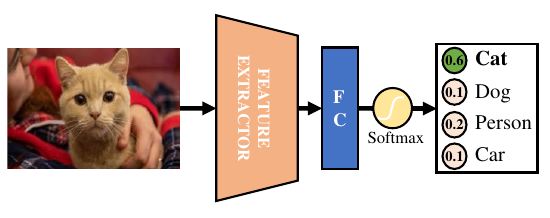}
    \caption{}
\end{subfigure}
\begin{subfigure}[b]{0.5\textwidth}
\centering
    \includegraphics[width=0.7\linewidth]{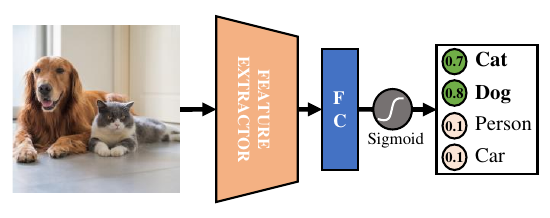}
    \caption{}
\end{subfigure}
\hfill
\caption{The work of ~\cite{daln} cannot be directly applied to MLIC due to the differences between the two tasks: (a) Single-label image classification uses a softmax activation function to convert the predicted logits into probabilities such that the sum of all class probabilities is equal to one; and (b) On the other hand, multi-label image classification uses sigmoid activation where each logit is scaled between $0$ and $1$, giving higher probability values for the objects present in an image.}
\label{fig:difference}
\end{figure}
With the latest advancements in deep learning, several MLIC methods~\cite{resnet,tresnet,asl, iml-gcn} have achieved remarkable performance on well-known datasets~\cite{coco,voc}. Nevertheless, the effectiveness of deep learning-based methods widely relies on the availability of annotated datasets. This requires costly and time-consuming efforts. As a result, given the limited number of labeled data, existing MLIC methods tend to have poor generalization capabilities to unseen domains. This problem is commonly known as \textit{domain-shift}, where a method trained on a \textit{source} dataset fails to generalize on a \textit{target} one belonging to a different domain. To overcome this issue, \textit{Unsupervised Domain Adaptation (UDA)}~\cite{dann,mada} can be an interesting strategy. The idea behind UDA is to leverage unlabeled data from the target dataset to reduce the gap between the source and the target domains.

In the literature, many works have been proposed for UDA in the context of single-label image classification~\cite{dann,mada,mcd,cada,mk-mmd}, while less efforts have been dedicated to proposing UDA methods that are suitable for MLIC. Inspired by the predominance of adversarial-based approaches in single-label image classification, few methods~\cite{da-maic,chest-xray,singh2023multi} have attempted to extend UDA to MLIC. Similar to~\cite{dann}, these adversarial approaches leverage a domain discriminator for implicitly reducing the domain gap. In particular, a min-max two-player game guides the generator to extract domain-invariant features that fool the discriminator. Nevertheless, this may come at the cost of decreasing their task-specific discriminative power, as highlighted in ~\cite{daln}. 

Chen et al.~\cite{daln} attempted to solve this problem in the context of single-label image classification by implicitly reusing the classifier as a discriminator. In particular, they considered the difference between inter-class and intra-class correlations of the classifier probability predictions as an adversarial critic. 
Nevertheless, the per-class prediction probabilities are not linearly dependent in the context of MLIC. This means that these probabilities are not constrained to sum up to one, as shown in Figure~\ref{fig:difference}. Hence, the approach of~\cite{daln} can only be naively generalized to MLIC by considering multiple binary classifiers, namely, one for each label. Therefore, a critic similar to the one in~\cite{daln} can be used by computing the correlations between the probability predictions of each binary classifier.  However, this is not optimal since the domain adaptation would be carried out for each label classifier separately, ignoring the correlations between the different labels. This is also experimentally confirmed in Section~\ref{sec:exp}.

\begin{figure}[!t]
\begin{minipage}[b]{.49\linewidth}
  \label{fig:intuition_source}
  \centering
  \centerline{\includegraphics[width=4.1cm]{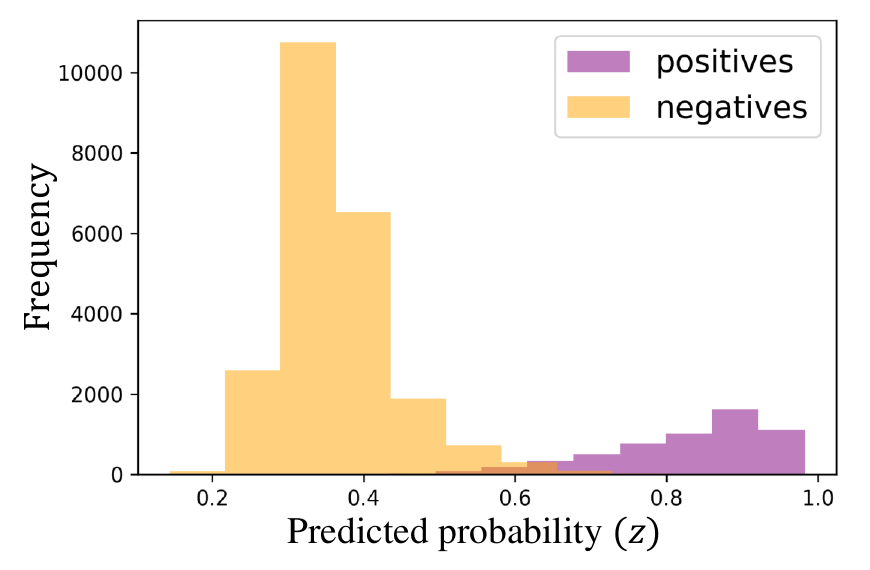}}
  \centerline{(a) Source $\rightarrow$ Source}\medskip
\end{minipage}
\begin{minipage}[b]{.49\linewidth}
\label{fig:intuition_target}
  \centering
  \centerline{\includegraphics[width=4.1cm]{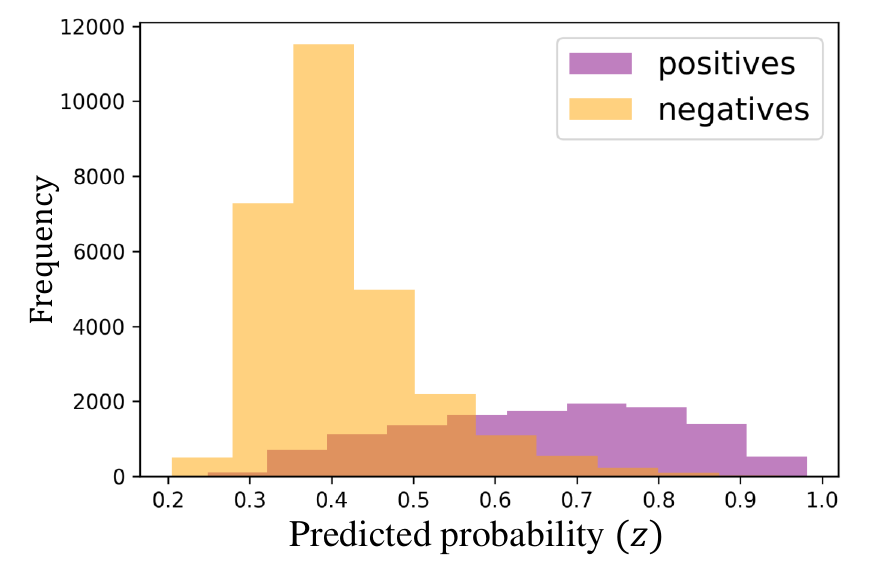}}
  \centerline{(b) Source $\rightarrow$ Target}\medskip
\end{minipage}
\hfill
\caption{Histogram of classifier predictions\protect\footnotemark[2]. Predicted probabilities using source-only trained classifier\protect\footnotemark[2] on: (a) source dataset\protect\footnotemark[3] $(\mathcal I_s)$, and (b) target dataset\protect\footnotemark[3] $(\mathcal I_t)$.}
\label{fig:intuition}
\end{figure}

In this paper, we introduce a discriminator-free adversarial UDA approach for MLIC based on a novel adversarial critic. As in~\cite{daln}, we propose to leverage the task-specific classifier for defining the adversarial critic. However, instead of relying on the prediction correlations, which is not suitable in the case of MLIC, we propose to cluster the probability predictions into two sets (one in the neighborhood of 0 and another one in the neighborhood of 1), estimate their respective distributions and define the critic as the distance between the estimated distributions from the source and target data.  This intuition comes from the fact that source data are usually more confidently classified (as positive or negative) than target ones, as illustrated in Figure~\ref{fig:intuition}. The same figure also highlights that the distribution of predictions can be modeled by two clusters; showing the interest of modelling the predictions with a bimodal distribution. Hence, we assume that the distribution shape of probability predictions can be used to implicitly discriminate source and target data. Concretely, we propose to fit a Gaussian Mixture Model (GMM) with two components on both the source and target predictions. A Fr\'echet distance~\cite{frechet} between the estimated pair of components is then used to define the proposed discrepancy measure. The experimental results show that the proposed approach outperforms state-of-the-art methods in terms of mean Average Precision (mAP) while significantly reducing the number of network parameters.

In summary, our contributions are:
\begin{itemize}
    \item A novel domain discrepancy for multi-label image classification based on the distribution of the task-specific classifier predictions;
    \item An effective and efficient adversarial unsupervised domain adaptation method for multi-label image classification. The proposed adversarial strategy does not require an additional discriminator, hence reducing the network size during training;
    \item An experimental quantitative and qualitative analysis on several benchmarks showing that the proposed method outperforms state-of-the-art works.
\end{itemize}

\footnotetext[2]{TResNet-M~\cite{tresnet} trained on UCM~\cite{ucm-ml} dataset.}
\footnotetext[3]{Source: UCM~\cite{ucm-ml} validation set (420 images), Target: AID~\cite{aid-ml} validation set (600 images).}

The rest of the paper is organized as follows. Section~\ref{sec:prob} formulates the problem of domain adaptation for multi-label image classification, and presents our intuition behind using the classifier as a critic. Section~\ref{sec:method} introduces the proposed approach termed DDA-MLIC. The experimental results are reported and discussed in Section~\ref{sec:exp}. Finally, Section~\ref{sec:conclusion} concludes this work and draws some perspectives.

\section{Problem Formulation and Motivation}
\label{sec:prob}

\subsection{Problem formulation}
Let $\mathcal D_s= (\mathcal I_s, \mathcal Y_s)$ and $\mathcal D_t= (\mathcal I_t, \mathcal Y_t)$ be the source and target datasets, respectively, with $P_s$ and $P_t$ being their respective probability distributions such that $P_s \neq P_t$. Let us assume that they are both composed of $N$ object labels. Note that $\mathcal I_s=\{\mathbf{I}_s^k\}_{k=1}^{n_s}$ and $\mathcal I_t=\{\mathbf{I}_t^k\}_{k=1}^{n_t}$  refer to the sets of $n_s$ source and $n_t$ target image samples, respectively, while $\mathcal Y_s=\{\mathbf{y}_s^k\}_{k=1}^{n_s}$ and $\mathcal Y_t=\{\mathbf{y}_t^k\}_{k=1}^{n_t}$ are their associated sets of labels. 

Let us denote by $\mathcal I$ the set of all images such that $\mathcal{I}~=~\mathcal{I}_s~\cup~\mathcal{I}_t$. Given an input image $\mathbf I \in \mathcal I $ with $\mathbf y\in \{0,1\}^N$ being its label, the goal of \textit{unsupervised domain adaptation} for \textit{multi-label image classification} is to estimate a function $f:\mathcal I \mapsto \{0,1\}^N$ such that,

\begin{equation}
    f(\mathbf I)=\mathbbm{1}_{f_c \circ f_g({\mathbf I})> \tau} = \mathbbm{1}_{\mathbf{Z}>\tau}=\mathbf y \ ,
\end{equation}

\noindent
where $f_g:~\mathcal{I}~\mapsto~\mathbb{R}^d$ extracts $d$-dimensional features, $f_c:~\mathbb R^d~\mapsto~[0,1]^N$ predicts the probability of object presence, $\mathbf{Z}=f_c\circ f_g (\mathbf I)\in [0,1]^N$ corresponds to the predicted probabilities, $\mathbbm{1}$ is an indicator function, $>$ is a comparative element-wise operator with respect to a chosen threshold  $\tau$. 
Note that only $\mathcal D_s$ and $\mathcal{I}_t$ are used for training.  In other words, the target dataset is assumed to be unlabeled.

To achieve this goal, some existing methods~\cite{da-maic} have adopted an adversarial strategy by considering an additional discriminator $f_d$ that differentiates between source and target data. Hence, the model is optimized using a classifier loss $\mathcal L_{c}$ such as the asymmetric loss (ASL)~\cite{asl} and an adversarial loss $\mathcal L_{adv}$ defined as,
\begin{equation}
\begin{split}
\label{eq:domain_loss}
    \mathcal L_{adv} = \mathbb{E}_{f_{g}(\mathbf I_s) \sim  \bar{P}_s}
 \log \frac{1}{f_d(f_g(\mathbf I_s))} + \\ \mathbb{E}_{f_{g}(\mathbf I_t) \sim \bar{P}_t}  \log \frac{1}{(1-f_d(f_g(\mathbf I_t))} \ ,
 \end{split}
\end{equation}
where $\bar{P}_s$ and $\bar{P}_t$ are the distributions of the learned features from source and target samples $\mathcal I_s$ and $\mathcal{I}_t$, respectively.

While the adversarial paradigm has shown great potential~\cite{da-maic}, the use of an additional discriminator $f_d$ which is decoupled from $f_c$ may lead to mode collapse as discussed in~\cite{daln}.  Inspired by the same work, we aim at addressing the following question -- \textit{Could we leverage the outputs of the task-specific classifier $f_c \circ f_g$ in the context of multi-label classification for implicitly discriminating the source and the target domains?}

\subsection{Motivation: domain discrimination using the distribution of the classifier output  }
\label{sec:critic}
The goal of MLIC is to identify the classes that are present in an image (\emph{i.e.}, \textit{positive labels}) and reject the ones that are not present (\emph{i.e.}, \textit{negative labels}). Hence, the classifier $f_c$ is expected to output high probability values for the positive labels and low probability values for the negative ones. Formally, let $z=\theta (f_c(f_g(\mathbf I))) =\theta(\mathbf{Z} )\sim \hat P$ be the random variable modelling the predicted probability of any class and $\hat P$ its probability distribution, with $\theta$ being a uniform random sampling function that returns the predicted probability of a randomly selected class. In general, a well-performing classifier is expected to classify confidently both negative and positive samples. Ideally, this would mean that the probability distribution $\hat P$ should be formed by two clusters with low variance in the neighborhood of $0$ and $1$, respectively denoted by $\mathcal C_0$ and $\mathcal C_1$. Hence, our hypothesis is that a drop in the classifier performance due to a domain shift can be reflected in $\hat P$. 

Let $z_s=\theta (f_c(f_g(\mathbf I_s)))$ $\sim \hat{P_s}$ and $z_t= \theta (f_c(f_g(\mathbf I_t))) \sim \hat P_t$ be the random variables modelling the predicted probability obtained from the source and target data and $\hat P_s$ and $\hat P_t$ be their distributions, respectively. Concretely, we propose to investigate whether the shift between the source and target domains is translated in  $\hat P_s$ and $\hat P_t$. If a clear difference is observed between $\hat P_s$ and $\hat P_t$,  this would mean that the classifier $f_c$ should be able to discriminate between source and target samples. Thus, this would allow the definition of a suitable critic directly from the classifier predictions.

To support our claim, we trained a model\protect\footnotemark[4] $f$
 using the labelled source data $\mathcal{D}_s$ without involving the target images\protect\footnotemark[5] $\mathcal{I}_t$. In Figure~\ref{fig:intuition}~(a), we visualize the histogram of the classifier probability outputs when the model is tested on the source domain. It can be clearly observed that the predicted probabilities on the source domain, denoted by $z_s$, can be grouped into two separate clusters.  Figure~\ref{fig:intuition}~(b) shows the same histogram when the model is tested on target samples. In contrast to the source domain, the classifier probability outputs, denoted by $z_t$, are more spread out in the target domain. In particular, the two clusters are less separable than in the source domain. This is due to the fact that the classifier $f_c$ benefited from the supervised training on the source domain, and as a result it gained an implicit discriminative ability between the source and target domains. 

Motivated by the observations discussed above, we propose to reuse the classifier to define a critic function based on $\hat P_s$ and $\hat P_t$. In what follows, we describe our approach including the probability distribution modelling ($\hat P_s$ and $\hat P_t$) and the adversarial strategy for domain adaptation.   

\footnotetext[4]{TResNet-M~\cite{tresnet} trained on UCM~\cite{ucm-ml} dataset.}
\footnotetext[5]{Source: UCM~\cite{ucm-ml} validation set (420 images), Target: AID~\cite{aid-ml} validation set (600 images).}

\section{An Implicit Multi-Label Domain Adaptation Adversarial Strategy}
\label{sec:method}
\begin{figure}[!t]
\centering
\begin{minipage}[b]{0.48\linewidth}
  \centering
\centerline{\includegraphics[width=4.1cm]{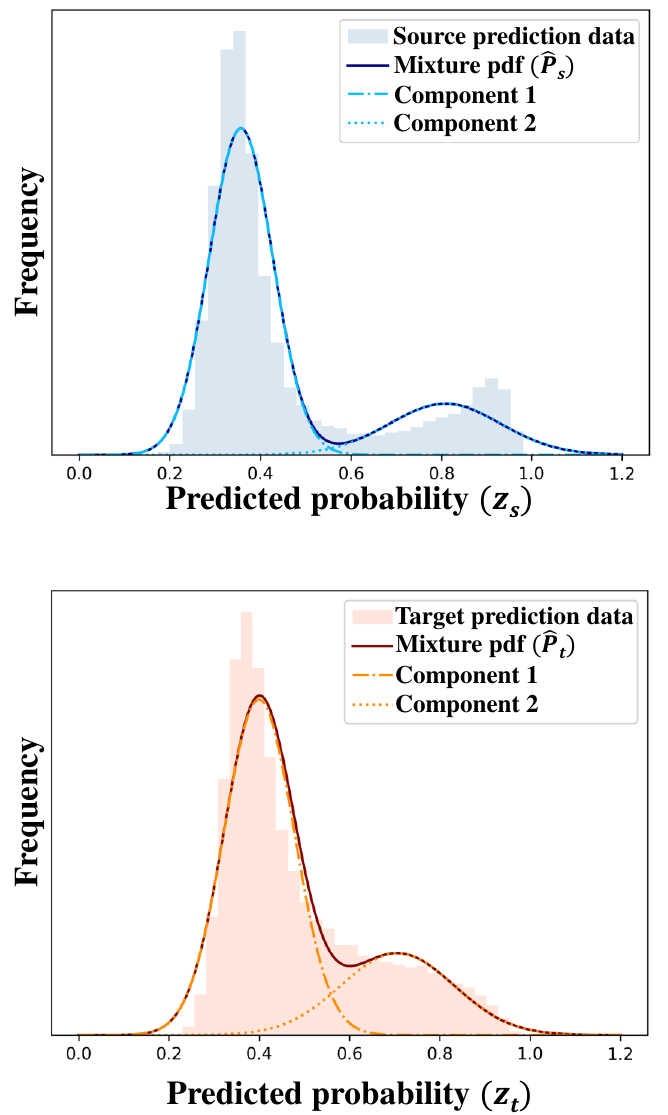}}
  \centerline{(a) GMM fitting.}
\end{minipage}
\begin{minipage}[b]{0.48\linewidth}
  \centering
  \centerline{\includegraphics[width=4.1cm]{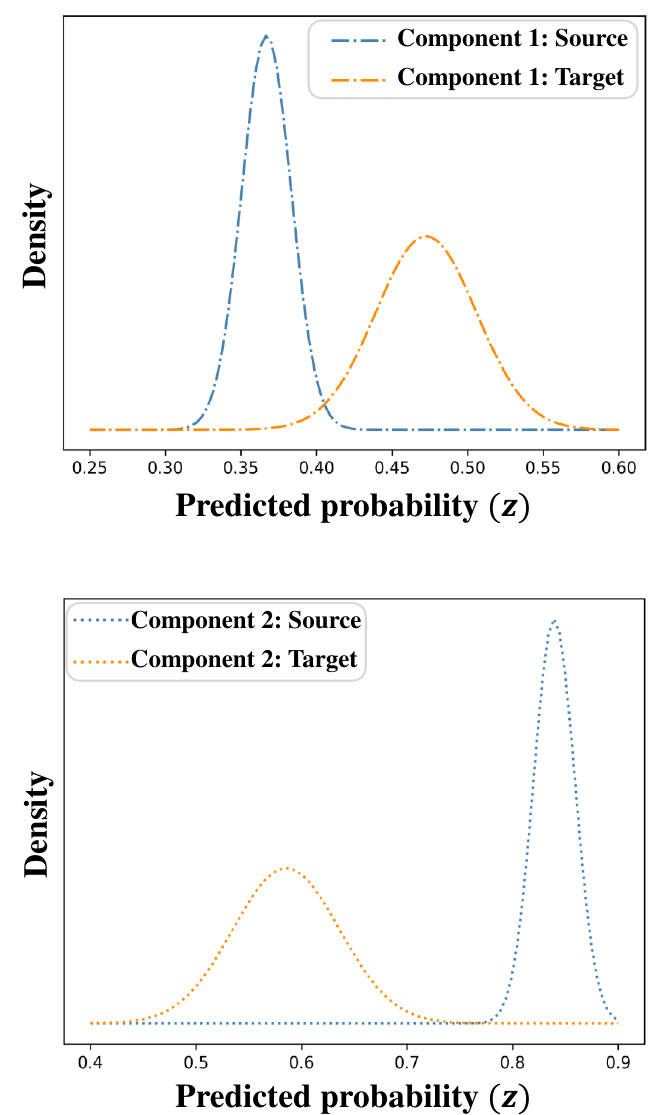}}
  \centerline{(b) Gaussians of components.}
\end{minipage}
\hfill
\caption{(a) The classifier\protect\footnotemark[6] predictions $z_s$ and $z_t$ for both source and target datasets\protect\footnotemark[7], respectively, can be grouped into two clusters. Hence, a two-component GMM can be fitted for both source ($\hat{P}_s$) and target ($\hat{P}_t$). While the first component is close to 0, the second is close to 1, (b) A component-wise comparison between source ($\hat P_s^1, \hat P_s^2$) and target ($\hat P_t^1, \hat P_t^2$) Gaussians of distributions extracted from the fitted GMM confirms that target predictions are likely to be farther from 0 and 1 with a higher standard deviation than the source.}

\label{fig:intro} 
\end{figure}

\begin{figure*}[!t]
  \centering
  \centerline{\includegraphics[width=0.9\linewidth]{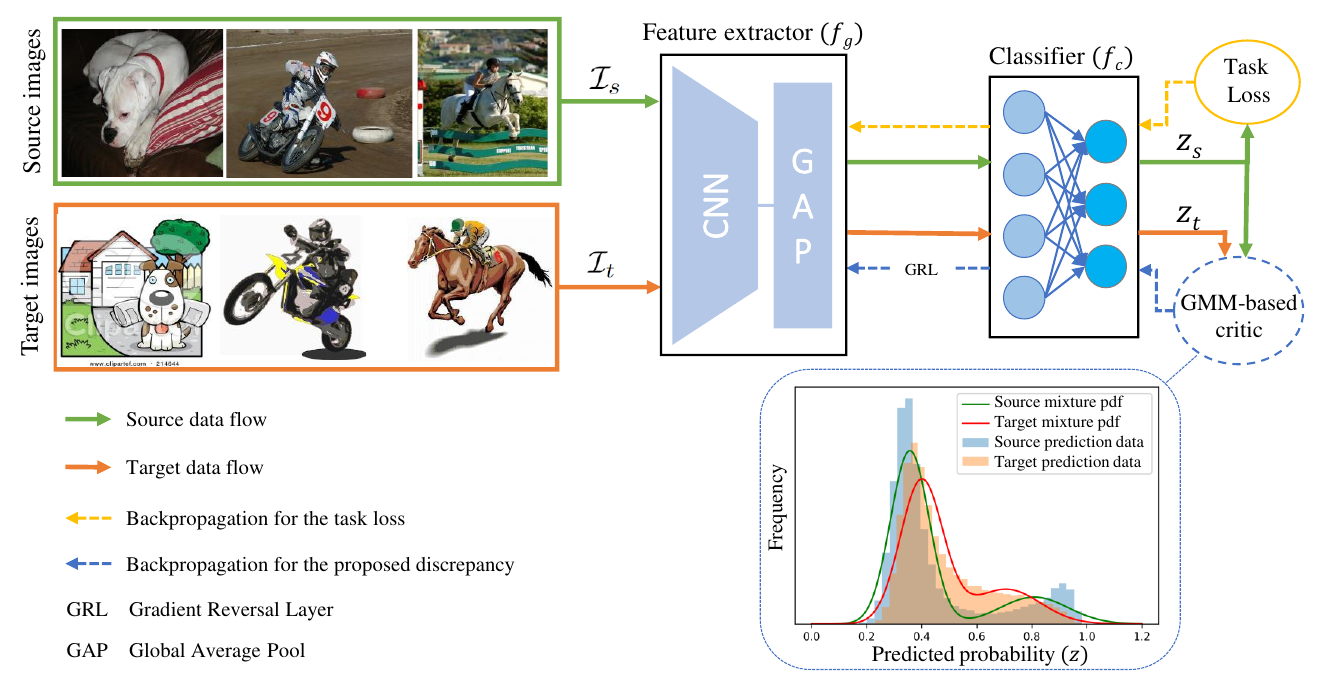}}
\caption{Overall proposed architecture of DDA-MLIC: The feature extractor ($f_g$) learns discriminative features from source and target images. The task classifier ($f_c$) performs two actions simultaneously: 1) it learns to classify source samples correctly using a supervised task loss, and 2) when used as a discriminator, it aims to minimize/maximize the proposed discrepancy between source and target predictions.}
\label{fig:arch}
\end{figure*}

As discussed in Section~\ref{sec:critic}, the classifier probability predictions are usually formed by two clusters with nearly Gaussian distributions. Consequently, as shown in Figure~\ref{fig:intro} (a), we propose to approximate the distributions $\hat P_s$ and $\hat P_t$ by a two-component Gaussian Mixture Model (GMM) as follows,

\begin{equation}
\label{eq:gmm_s}
   \hat P_s(z_s) \approx \sum_{i=1}^2 \pi_i^s \mathcal{N}(z_s|\mu_i^s, \sigma_i^s) \ ,
\end{equation} 
and, 
\begin{equation}
\label{eq:gmm_t}
   \hat P_t(z_t) \approx \sum_{i=1}^2 \pi_i^t \mathcal{N}(z_t|\mu_i^t, \sigma_i^t) \ ,
\end{equation} 

\noindent where $\mathcal{N}(z_t|\mu_i^t, \sigma_i^t)$ denotes the $i$-th Gaussian distribution, with the mean  $\mu_i^t$ and the variance $\sigma_i^t$, fitted on the target predicted probabilities $z_t$ and $\pi_i^t$ its mixture weight such that $\pi_1^t+\pi_2^t=1$. Similarly,  $\mathcal{N}(z_s|\mu_i^s, \sigma_i^s)$ denotes the $i$-th Gaussian distribution, with the mean $\mu_i^s$ and the variance $\sigma_i^s$, fitted on the source predicted probabilities $z_s$ and $\pi_i^s$ its mixture weight such that $\pi_1^s+\pi_2^s=1$.

An Expectation-Maximization (EM) algorithm is used to estimate the GMM parameters. In both source and target domains, we assume that the first component of the GMM corresponds to the cluster  $\mathcal C_0$ (with a mean close to 0), while the second corresponds to $\mathcal C_1$ (with a mean close to $1$). 

However, due to a large number of negative predictions as compared to positive ones, the component $\mathcal C_0$ tends to be more dominant. In fact, in a given image, only few objects are usually present from the total number of classes. To alleviate this phenomenon, we propose to extract two Gaussian components from the source and target GMM, ignoring the estimated weights illustrated in Figure~\ref{fig:intro} (b).

Hence, we propose to redefine the adversarial loss $\mathcal{L}_{adv}$ by computing a Fr\'echet distance $d_F$~\cite{frechet} between each pair of source and target components from a given cluster as follows. 

\begin{equation}
\label{eq:loss-adv}
    \mathcal{L}_{adv} = \sum_{i=1}^2  \alpha_i d_\mathrm{F}(\mathcal{N}(z_t|\mu_i^t, \sigma_i^t),\mathcal{N}(z_s|\mu_i^s, \sigma_i^s)) \ ,
\end{equation}

\noindent with $\alpha_i$ weights that are empirically fixed. 
Since the computed distributions are univariate Gaussians, the Fréchet distance between two distributions, also called the 2-Wasserstein (2W) distance, is chosen as it can be explicitly  computed as follows,

\begin{equation}
    d^2_\mathrm{F}(\mathcal N(z_1|\mu_1, \sigma_1),\mathcal N(z_2|\mu_2, \sigma_2)) = 
    (\mu_1-\mu_2)^2+(\sigma_1-\sigma_2)^2,
\end{equation}
where $\mathcal N(z_1|\mu_1, \sigma_1)$ and $ \mathcal N(z_2|\mu_2, \sigma_2)$ are two Gaussians with a mean of $\mu_1$ and $\mu_2$ and a standard deviation of $\sigma_1$ and $\sigma_2$, respectively. In addition, compared to the commonly used 1-Wasserstein (1W) distance, it considers second-order moments. Finally, in~\cite{arjovsky2017wasserstein}, the 2W distance has been demonstrated to have nicer properties e.g., continuity and differentiability,  for optimizing neural networks as compared to other divergences and distances between two distributions such as the Kullback-Leibler  (KL) divergence and the Jensen-Shannon (JS) divergence. The relevance of the 2W distance is further discussed in Section~\ref{sec:distance}. 

The overall architecture of the proposed method is shown in Figure~\ref{fig:arch}. Similar to~\cite{daln}, our network consists of a feature extractor $f_g$ that aims to extract discriminative image features from source $\mathcal{D}_s$ and target $\mathcal{I}_t$ datasets and a classifier $f_c$ that simultaneously performs the classification and discriminates between source and target features by minimizing the proposed adversarial loss $\mathcal{L}_{adv}$. A Gradient Reversal Layer (GRL) between $f_g$ and $f_c$ enforces the feature extractor to fool the classifier when acting as a discriminator, hence implicitly learning domain-invariant features.

\footnotetext[6]{TResNet-M~\cite{tresnet} trained on UCM~\cite{ucm-ml} dataset.}
\footnotetext[7]{Source: UCM~\cite{ucm-ml} validation set (420 images), Target: AID~\cite{aid-ml} validation set (600 images).}

\section{Experiments}
\label{sec:exp}

\subsection{Datasets}
In our experiments, different types of domain gaps are considered. Due to the limited availability of multi-label domain adaptation datasets, we convert several object detection and semantic segmentation datasets for the task of MLIC.

\paragraph{Cross-sensor domain shift}
Similar to~\cite{da-maic}, we use three multi-label aerial image datasets that have been captured using different sensors resulting in different resolutions, pixel densities and altitudes, namely: 1) \textbf{AID}~\cite{aid-ml} multi-label dataset was created from the original multi-class AID dataset~\cite{aid} by labelling $3000$ aerial images, including $2400$ for training and $600$ for testing, with a total of $17$ categories. 2) \textbf{UCM}~\cite{ucm-ml} multi-label dataset was recreated from the original multi-class classification dataset~\cite{ucm} with a total of $2100$ image samples containing the same $17$ object labels as AID. We randomly split the dataset into training and testing sets with 2674 and 668 image samples, respectively. 3) \textbf{DFC}~\cite{dfc} multi-label dataset provides $3342$ high resolution images with  training and testing splits of, respectively, $2674$ and $668$ samples labelled from a total of $8$ categories. In our experiments, the $6$ common categories between DFC and the other two benchmarks are used.

\begin{table*}[!t]
\caption{Cross-sensor domain shift: Comparison with the state-of-the-art in terms of number of model parameters (in millions), and \% scores of mAP, per-class averages (CP, CR, CF1) and overall averages (OP, OR, OF1) for aerial image datasets. Two settings are considered, \emph{i.e.}, AID $\rightarrow$ UCM and UCM $\rightarrow$ AID. Best results are highlighted in \textbf{bold}.}
\label{table:aid_ucm_results_both}
\resizebox{\textwidth}{!}{%
\begin{tabular}{|l|l|c|ccccccc|ccccccc|}
\hline
\multirow{2}{*}{\textbf{Type}} & \multirow{2}{*}{\textbf{Method}} & \multirow{2}{*}{\textbf{\# params}} & \multicolumn{7}{c}{\textbf{AID $\rightarrow$ UCM}}                                                           & \multicolumn{7}{|c|}{\textbf{UCM $\rightarrow$ AID}}                                                           \\
\cline{4-17}
                               &                                  &                                     & mAP           & P\_C          & R\_C          & F\_C          & P\_O          & R\_O          & F\_O          & mAP           & P\_C          & R\_C          & F\_C          & P\_O          & R\_O          & F\_O          \\
                               \hline
\multirow{4}{*}{MLIC}          & ResNet101~\cite{resnet}                        & 42.5                                & 57.5          & \textbf{60.0} & 47.5          & 47.0          & 69.1          & 71.5          & \textbf{70.3} & 51.7          & 50.6          & 29.6          & 33.9          & 88.0          & 48.5          & 62.5          \\
                               & ML-GCN~\cite{ml-gcn}                           & 44.9                                & 53.7          & 55.3          & 44.3          & 45.9          & 70.2          & 68.7          & 69.4          & 51.3          & 50.1          & 29.9          & 34.0          & 88.0          & 49.7          & 63.6          \\
                               & ML-AGCN~\cite{ml-agcn}                          & 36.6                                & 55.2          & 36.6          & \textbf{64.9} & 45.1          & 45.0          & \textbf{88.1} & 59.6          & 52.1          & 48.2          & \textbf{47.4} & \textbf{42.9} & 77.1          & \textbf{79.8} & \textbf{78.4} \\
                               & ASL   (TResNetM)~\cite{asl}                 & \textbf{29.4}                       & 55.4          & 48.7          & 52.8          & 47.1          & 58.7          & 79.1          & 67.4          & 54.1          & 54.5 & 40.2          & 41.9          & 85.4          & 65.1          & 73.9          \\
                               \hline
\multirow{2}{*}{Disc.-based}   & DANN (TResNetM+ASL)~\cite{dann}              & \textbf{29.4}                       & 52.5          & 59.1          & 31.6          & 36.3          & \textbf{70.9} & 53.7          & 61.1          & 51.6          & 52.1          & 23.2          & 27.9          & 83.2          & 27.8          & 41.7          \\
                               & DA-MAIC   (TResNetM+ASL)~\cite{da-maic}         & 36.6                                & 54.4          & 55.3          & 37.5          & 38.6          & 68.0          & 67.9          & 67.9          & 50.5          & 51.8          & 22.9          & 29.0          & 91.6 & 35.2          & 50.8          \\
                               \hline
\multirow{3}{*}{Disc.-free}    & DALN (TResNetM+ASL)~\cite{daln}              & \textbf{29.4}                       & 53.1          & 53.3          & 32.4          & 36.7          & 69.2          & 53.9          & 60.6          & 53.2          & 52.2          & 29.3          & 32.7          & 82.0          & 41.2          & 54.8          \\
\cline{2-17}
                               & \textbf{DDA-MLIC   (ours)}       & \textbf{29.4}                       & \textbf{63.2} & 52.5 & 63.7 & \textbf{55.1} & 59.4 & 82.8 & 69.2         & \textbf{54.9} & 53.9          & 30.4          & 35.5          & 84.6          & 41.0          & 55.3         \\
                               \hline
\end{tabular}}
\end{table*}

\begin{table*}[!t]
\caption{Cross-sensor domain shift: Comparison with the state-of-the-art in terms of number of model parameters (in millions), and \% scores of mAP, per-class averages (CP, CR, CF1) and overall averages (OP, OR, OF1) for aerial image datasets. Two settings are considered, \emph{i.e.}, AID $\rightarrow$ DFC and UCM $\rightarrow$ DFC. Best results are highlighted in \textbf{bold}.}
\label{table:dfc_all}
\resizebox{\textwidth}{!}{%
\begin{tabular}{|l|l|c|ccccccc|ccccccc|}
\hline
                                &                                   &                                      & \multicolumn{7}{c}{\textbf{AID $\rightarrow$ DFC}}                                                                          & \multicolumn{7}{|c|}{\textbf{UCM $\rightarrow$ DFC}}                                                                          \\
                                \cline{4-17}
\multirow{-2}{*}{\textbf{Type}} & \multirow{-2}{*}{\textbf{Method}} & \multirow{-2}{*}{\textbf{\# params}} & mAP                          & P\_C          & R\_C          & F\_C          & P\_O          & R\_O          & F\_O          & mAP                          & P\_C          & R\_C          & F\_C          & P\_O          & R\_O          & F\_O          \\
\hline
                                & ResNet101~\cite{resnet}                         & 42.5                                 & 56.9                         & 52.9          & 61.5          & 48.7          & 46.1          & 63.7          & 53.5          & 66.4                         & 74.4          & 31.2          & 36.9          & 67.2          & 37.2          & 47.9          \\
                                & ML-GCN~\cite{ml-gcn}                            & 44.9                                 & 58.9                & \textbf{56.7} & 57.9          & 45.8          & 45.7          & 65.0          & 53.7          & 64.6                         & 72.4          & 32.0          & 35.6          & 64.4          & 38.9          & 48.5          \\
                                & ML-AGCN~\cite{ml-agcn}                           & 36.6                                 & 51.6                         & 41.5          & \textbf{83.8} & 52.3 & 40.2          & \textbf{88.7} & 55.3          & 70.3                & 68.4          & \textbf{56.1} & 47.8 & 53.8          & \textbf{58.5} & 56.0 \\
\multirow{-4}{*}{MLIC}          & ASL   (TResNetM)~\cite{asl}                  & \textbf{29.4}                        & 56.1                         & 49.6          & 68.4          & 49.9          & 43.5          & 74.1          & 54.8          & 68.9                         & 66.3          & 53.1          & 44.0          & 52.6          & 57.0          & 54.7          \\
\hline
                                & DANN (TResNetM+ASL)~\cite{dann}               & \textbf{29.4}                        & 43.0                         & 40.7          & 13.6          & 19.3          & 46.0          & 15.6          & 23.3          & 64.1                         & 77.3          & 22.6          & 30.1          & 68.6          & 26.5          & 38.2          \\
\multirow{-2}{*}{Disc.-based}   & DA-MAIC   (TResNetM+ASL)~\cite{da-maic}          & 36.6                                 & 55.4                         & 49.8          & 60.4          & 44.7          & 47.3          & 64.1          & 54.4          & 65.8                         & 71.4          & 39.3          & 39.7          & 59.9          & 44.6          & 51.1          \\
\hline
                                & DALN (TResNetM+ASL)~\cite{daln}               & \textbf{29.4}                        & 44.7                         & 43.7          & 23.8          & 27.6          & \textbf{48.9}          & 27.4          & 35.1          & 65.6                         & \textbf{82.6} & 21.3          & 32.0          & \textbf{75.2} & 22.1          & 34.1          \\
                                \cline{2-17}
\multirow{-2}{*}{Disc.-free}    
& \textbf{DDA-MLIC   (ours)}        & \textbf{29.4}                        & \textbf{62.1} & 47.6 & 75.5 & \textbf{55.3} & 48.9 & 76.2 & \textbf{59.6} & \textbf{70.6} & 67.2 & 55.7 & \textbf{49.3} & 55.0 & 58.4 & \textbf{56.6}         \\
\hline
\end{tabular}}
\end{table*}

\paragraph{Sim2real domain shift}
We use the following two datasets to investigate the domain gap between real and synthetic scene understanding images. 1) \textbf{PASCAL-VOC}\cite{voc} is one of the most widely used real image datasets for MLIC with more than $10K$ image samples. It covers $20$ object categories. The training and testing sets contain $5011$ and $4952$ image samples, respectively. 2) \textbf{Clipart1k}~\cite{clipart} provides 1000 synthetic clipart image samples, annotated with 20 object labels, similar to VOC. Since it is proposed for the task of object detection, we make use of the category labels for bounding boxes to create a multi-label version. Half of the data are used for training and the rest is used for testing. 




\paragraph{Cross-weather domain shift}
In order to study the domain shift caused by different weather conditions, two widely used urban street datasets have been used, namely:
1) \textbf{Cityscapes}~\cite{cityscapes} which is introduced for the task of semantic image segmentation and consists of $5000$ real images captured in daytime. 2) \textbf{Foggy-cityscapes}~\cite{foggy} which is a synthesized version of Cityscapes where an artificial fog is introduced. We generate a multi-label version of these datasets for the task of MLIC considering only $11$ categories out of the original $19$ to avoid including the objects that appear in all the images. 
\subsection{Implementation details}
The proposed DDA-MLIC makes use of the TResNet-M~\cite{tresnet} as a backbone and the Asymmetric Loss (ASL)~\cite{asl} as the task loss. All the methods are trained using the Adam optimizer with a cosine decayed maximum learning rate of $10^{-3}$. For all the experiments, we make use  of NVIDIA TITAN V with a batch size of $64$ for a total of $25$ epochs or until convergence. The input image resolution has been fixed to  $224\times224$..
\subsection{Baselines}
\label{sec:baselines}
To evaluate the proposed approach, we consider standard MLIC approaches, namely, ResNet~\cite{resnet}, ML-GCN~\cite{ml-gcn}, ML-AGCN~\cite{ml-agcn} and ASL~\cite{asl} as well the recently introduced DA method for MLIC approach called DA-MAIC~\cite{da-maic}. Note that the standard MLIC approaches are trained on source-only datasets, hence do not incorporate any domain adaptation strategy. In addition, given that the problem of DA for MLIC is under-explored, we propose to adapt two additional DA methods for single-label classification to MLIC. In particular, the existing discriminator-based and discriminator-free adversarial DA approaches i.e., DANN~\cite{dann} and DALN~\cite{daln} are considered by adopting the following changes. The original cross entropy loss in both DANN and DALN is replaced with the Asymmetric Loss (ASL)~\cite{asl}. Additionally, we propose to convert the multi-label output of the classifier in DALN to multiple binary predictions before applying the Nuclear Wasserstein Discrepancy (NWD)~\cite{daln}. Moreover, for a fair comparison, we replace the CNN backbone from the conventional ResNet101 to the same backbone as ours, namely TResNet-M ~\cite{tresnet}, for the three DA baselines.

\subsection{Experimental settings}
In our experiments, we report the number of model parameters, mean Average Precision (mAP), average per-Class Precision (CP), average per-Class Recall (CR), average per-Class F1-score (CF1), average Overall Precision (OP), average overall recall (OR) and average Overall F1-score (OF1). Given the seven considered datasets, \emph{i.e.}, AID, UCM, DFC, VOC, Clipart, Cityscapes and Foggycityscapes, seven experimental settings are considered, \emph{i.e.}, AID $\rightarrow$ UCM, UCM $\rightarrow$ AID, AID $\rightarrow$ DFC, UCM $\rightarrow$ DFC, VOC $\rightarrow$ Clipart, Clipart $\rightarrow$ VOC, Cityscapes $\rightarrow$ Foggy. For instance, AID $\rightarrow$ UCM indicates that during the training AID is fixed as the source dataset while UCM is considered as the target one. The results are reported on the testing set of the target dataset. 

\begin{table*}[!t]
\caption{Sim2Real domain shift: Comparison with the state-of-the-art in terms of number of model parameters (in millions), and \% scores for mAP, per-class averages (CP, CR, CF1) and overall averages (OP, OR, OF1) for scene understanding datasets. Two settings are considered, \emph{i.e.}, VOC $\rightarrow$ Clipart and Clipart $\rightarrow$ VOC. Best results are highlighted in \textbf{bold}.}
\label{table:voc_clipart}
\resizebox{0.95\textwidth}{!}{%
\begin{tabular}{|l|l|c|ccccccc|ccccccc|}
\hline
\multirow{2}{*}{\textbf{Type}} & \multirow{2}{*}{\textbf{Method}} & \multirow{2}{*}{\textbf{\# params}} & \multicolumn{7}{c}{\textbf{VOC $\rightarrow$ Clipart}}                                                       & \multicolumn{7}{|c|}{\textbf{Clipart $\rightarrow$ VOC}}                                                       \\
\cline{4-17}
                               &                                  &                                     & mAP           & P\_C          & R\_C          & F\_C          & P\_O          & R\_O          & F\_O          & mAP           & P\_C          & R\_C          & F\_C          & P\_O          & R\_O          & F\_O          \\
                               \hline
\multirow{4}{*}{MLIC}          & ResNet101~\cite{resnet}                        & 42.5                                & 38.0          & 64.8          & 14.3          & 22.5          & 82.3          & 18.3          & 29.9          & 50.1          & 66.2          & 17.5          & 25.5          & 83.9          & 29.6          & 43.7          \\
                               & ML-GCN~\cite{ml-gcn}                           & 44.9                                & 43.5          & 62.5          & 20.3          & 28.4          & 86.6          & 27.8          & 42.1          & 43.1          & 57.9          & 21.0          & 26.8          & 73.5          & 30.6          & 43.2          \\
                               & ML-AGCN~\cite{ml-agcn}                          & 36.6                                & 53.7          & 75.5          & 35.5          & 44.4          & 79.1          & 39.9          & 53.1          & 38.0          & 45.5          & 25.1          & 28.2          & 61.8          & 36.6          & 45.9          \\
                               & ASL   (TResNetM)~\cite{asl}                 & \textbf{29.4}                       & 56.8          & 72.0          & 38.5          & 47.6          & 82.8          & 45.7          & 58.9          & 64.2          & 69.0          & 30.7          & 37.3          & 80.0          & 45.7          & 58.2          \\
                               \hline
\multirow{2}{*}{Disc.-based}   & DANN (TResNetM+ASL)~\cite{dann}              & \textbf{29.4}                       & 47.0          & 77.0          & 22.0          & 32.5          & 86.8          & 23.6          & 37.1          & 67.0          & 76.8          & 23.3          & 32.6          & \textbf{93.1} & 20.4          & 33.4          \\
                               & DA-MAIC   (TResNetM+ASL)~\cite{da-maic}         & 36.6                                & \textbf{62.3} & 77.4          & \textbf{42.6} & \textbf{51.6} & 83.1          & \textbf{51.0} & \textbf{63.2} & 74.3          & 84.5          & \textbf{53.9} & \textbf{63.0} & 83.7          & \textbf{57.7} & \textbf{68.3} \\
                               \hline
\multirow{2}{*}{Disc.-free}    & DALN (TResNetM+ASL)~\cite{daln}              & \textbf{29.4}                       & 45.0          & 82.2          & 21.4          & 32.6          & \textbf{92.0} & 22.7          & 36.4          & 66.7          & 78.3          & 22.2          & 31.7          & 90.8          & 18.0          & 30.0          \\
\cline{2-17}
                               & \textbf{DDA-MLIC   (ours)}       & \textbf{29.4}                       & 61.4          & \textbf{84.7} & 28.1          & 39.4          & 90.9          & 33.3          & 48.8          & \textbf{77.0} & \textbf{86.9} & 29.3          & 38.2          & 88.4          & 35.3          & 50.4         \\                              
                               \hline
\end{tabular}}
\end{table*}

\begin{table*}[!t]
\caption{Cross-weather domain shift: Comparison with the state-of-the-art in terms of number of model parameters (in millions), and \% scores of mAP, per-class averages (CP, CR, CF1) and overall averages (OP, OR, OF1) for urban street datasets. Cityscapes $\rightarrow$ Foggy is the setting that is considered. Best results are highlighted in \textbf{bold}.}
\label{table:coco_city}
\centering
\resizebox{0.65\textwidth}{!}{%
\begin{tabular}{|l|l|c|ccccccc|}
\hline
\multirow{2}{*}{\textbf{Type}} & \multirow{2}{*}{\textbf{Method}} & \multirow{2}{*}{\textbf{\# params}} & \multicolumn{7}{c|}{\textbf{Cityscapes $\rightarrow$ Foggy}}                                                        \\
\cline{4-10}
                               &                                  &                                     & mAP           & P\_C          & R\_C          & F\_C          & P\_O          & R\_O          & F\_O          \\
                               \hline
\multirow{4}{*}{MLIC}          & ResNet101~\cite{resnet}                        & 42.5                                & 58.2          & 53.6          & 27.8          & 32.2          & \textbf{93.2} & 48.3          & 63.7         \\ 
& ML-GCN~\cite{ml-gcn}  & 44.9 & 56.6 & 56.1 & 34.6 & 38.8 & 89.4 & 56.9          & 69.6          \\
& ML-AGCN~\cite{ml-agcn} & 36.6 & 60.7 & 60.1 & 48.3 & 50.9 & 81.7 & \textbf{71.2} & \textbf{76.1} \\
                               & ASL   (TResNetM)~\cite{asl}                 & \textbf{29.4}                       & 61.3          & 66.7          & \textbf{50.8} & \textbf{53.8} & 79.2          & 70.5 & 74.6          \\
                               \hline
\multirow{2}{*}{Disc.-based}   & DANN (TResNetM+ASL)~\cite{dann}              & \textbf{29.4}                       & 53.5          & 50.6          & 12.5          & 18.6          & 89.5          & 21.8          & 35.1          \\
                               & DA-MAIC   (TResNetM+ASL)~\cite{da-maic}         & 36.6                                & 61.9          & 70.7          & 37.2          & 42.7          & 90.2          & 59.6          & 71.7          \\
                               \hline
\multirow{2}{*}{Disc.-free}    & DALN (TResNetM+ASL)~\cite{daln}              & \textbf{29.4}                       & 54.8          & 56.8          & 19.5          & 25.4          & 90.2          & 33.8          & 49.2          \\
\cline{2-10}
                               & \textbf{DDA-MLIC   (ours)}                & \textbf{29.4}                       & \textbf{62.3} & \textbf{73.7} & 45.7          & 48.9          & 84.1          & 69.3          & 76.0\\
                               \hline
\end{tabular}}
\end{table*}
\subsection{Results}
\subsubsection{Quantitative analysis}

Table~\ref{table:aid_ucm_results_both}, Table~\ref{table:dfc_all}, Table~\ref{table:voc_clipart} and Table~\ref{table:coco_city} quantitatively compare the proposed approach to state-of-the-art methods.  It can be seen that our model requires equal or fewer number of parameters than other state-of-the-art works, with a total number of $29.4$ million parameters. We achieve the best performance in terms of mAP for AID $\rightarrow$ UCM, UCM $\rightarrow$ AID, AID $\rightarrow$ DFC, UCM $\rightarrow$ DFC, Clipart $\rightarrow$ VOC and Cityscapes $\rightarrow$ Foggy. 

The first $4$ rows of Table~\ref{table:aid_ucm_results_both}, Table~\ref{table:dfc_all}, Table~\ref{table:voc_clipart} and Table~\ref{table:coco_city} report the obtained results using different methods of MLIC without DA~\cite{resnet,ml-gcn,ml-agcn,asl}. It can be observed that our method consistently outperforms all these methods in all settings (cross-sensor, sim2Real, and cross-weather) in terms of mAP showing the effectiveness of the proposed DA method for MLIC. 

Furthermore, the results reported in the $5^{th}$ and $6^{th}$ rows of Table~\ref{table:aid_ucm_results_both}, Table~\ref{table:dfc_all}, Table~\ref{table:voc_clipart}, Table~\ref{table:coco_city} show that the proposed discriminator-free DA method clearly outperforms discriminator-based DA approaches for MLIC~\cite{dann,da-maic} on cross-sensor and cross-weather domain shift settings in terms of mAP. This observation was not consistent for sim2Ream domain shift, where our approach recorded an mAP improvement of $2.7\%$ over other discriminator-based approaches on Clipart~$\rightarrow$~VOC setting, but was slightly outperformed with $0.9\%$ in terms of mAP by DA-MAIC~\cite{da-maic} on VOC $\rightarrow$ Clipart setting.

Finally, we compare our method to the discriminator-free method proposed in~\cite{daln} for single-label DA and adapted to MLIC as stated in Section~\ref{sec:baselines}. Unsurprisingly, our method consistently outperforms the adapted version of DALN for MLIC on all settings, reaching an improvement of more than $17\%$ in terms of mAP on the AID~$\rightarrow$~DFC scheme.

\subsubsection{Qualitative analysis}
\begin{figure*}[!t]
 \centering
 \centerline{\includegraphics[width=0.9\linewidth]{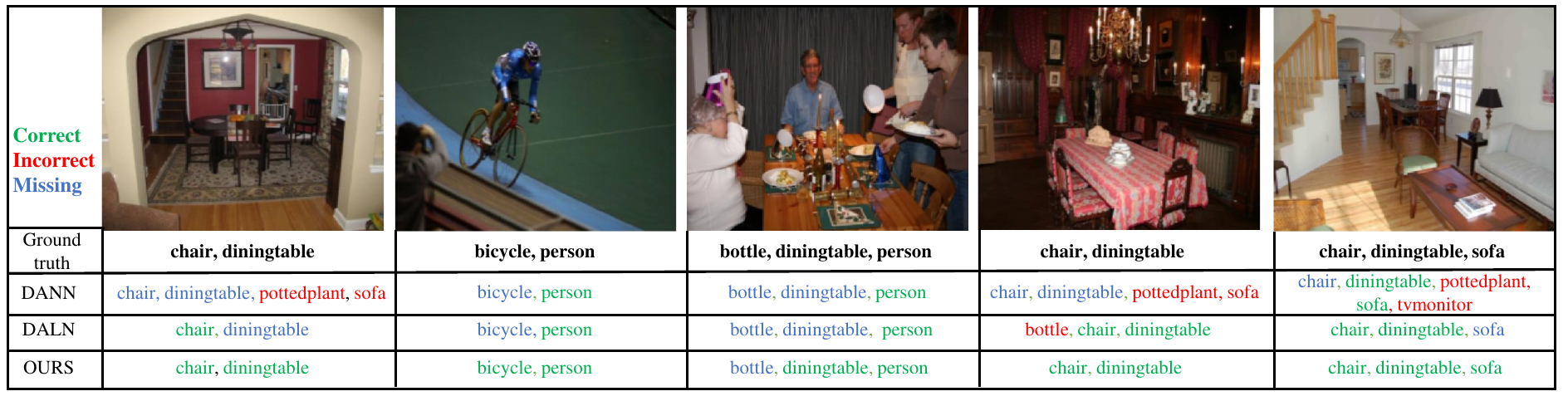}}
\caption{Qualitative analysis: Comparison of the proposed DDA-MLIC (OURS) with DANN~\cite{dann}  and DALN~\cite{daln} in terms of multi-label predictions on Clipart $\rightarrow$ VOC.}
\label{fig:qual_ml}
\end{figure*}

The proposed DDA-MLIC is qualitatively compared with DANN~\cite{dann} and DALN~\cite{daln} in Figure~\ref{fig:qual_ml} for the Clipart $\rightarrow$ VOC setting. The top row shows input images with their respective ground truth. The next three rows show the correct, incorrect,  and missing predictions, in green, red, and blue, respectively. It can be noted that the proposed approach correctly predicts the labels of the five image samples, in contrast to other DA methods that are failing in some cases. 

\subsection{Ablation study}
The results of the ablation study are shown in Table~\ref{table:ablation}.  We report the obtained mAP for the following settings, \emph{i.e.}, AID $\rightarrow$ UCM, UCM $\rightarrow$ AID, UCM $\rightarrow$ AID, UCM $\rightarrow$ DFC, VOC $\rightarrow$ Clipart and Clipart $\rightarrow$ VOC. The first row shows the mAP obtained in the absence of any domain adaptation strategy. The second row includes the scores obtained when adopting an adversarial domain adaptation approach using a standard domain discriminator. The third and last rows show the obtained results when using the proposed approach using a 2-Wasserstein distance.  It can be clearly seen that by leveraging the classifier as a discriminator, the classification performance is generally improved in the presence of a domain shift.

\begin{table*}[!t]
\centering
\caption{Ablation study (w/o: without, w/: with). The reported \% scores are mAP.}
\label{table:ablation}
\resizebox{0.7\textwidth}{!}{%
\begin{tabular}{|l|c|c|c|c|c|c|}
\hline
\textbf{Methods} & \textbf{AID$\rightarrow$UCM} & \textbf{UCM$\rightarrow$AID} & \textbf{AID$\rightarrow$DFC} & \textbf{UCM$\rightarrow$DFC} & \textbf{VOC$\rightarrow$Clipart} & \textbf{Clipart$\rightarrow$VOC} \\
\hline
\textbf{Ours}    & 63.24                           & 54.87                           & 62.13                           & 70.64                           & 61.44                               & 76.96                               \\
\hline
Ours w/o DA      & 55.45 \textbf{(-7.79)}                           & 54.12 \textbf{(-0.75)}                          & 56.09 \textbf{(-6.04)}                          & 68.91 \textbf{(-1.73)}                          & 56.78 \textbf{(-4.66)}                              & 64.15 \textbf{(-12.81)}                              \\
\hline
Ours w/ Discr.   & 52.54 \textbf{(-10.70)}                          & 51.60 \textbf{(-3.27)}                          & 51.60 \textbf{(-10.53)}                          & 64.06 \textbf{(-6.58) }                         & 46.97 \textbf{(-14.47)}                              & 67.03 \textbf{(-9.93)}                               \\
\hline
\end{tabular}}
\end{table*}

\begin{table*}[!t]
\caption{mAP comparison of using KL-divergence and 1-Wasserstein (1W) distance as discrepancy for domain alignment.}
\label{table:kl}
\centering
\resizebox{0.7\textwidth}{!}{%
\begin{tabular}{|l|c|c|c|c|c|c|}
\hline
\textbf{Methods}        & \textbf{AID$\rightarrow$UCM} & \textbf{UCM$\rightarrow$AID} & \textbf{AID$\rightarrow$DFC} & \textbf{UCM$\rightarrow$DFC} & \textbf{VOC$\rightarrow$Clipart} & \textbf{Clipart$\rightarrow$VOC} \\
\hline
\textbf{Ours} & 63.24                           & 54.90                           & 62.13                           & 70.64                           & 61.44                               & 76.96                               \\
\hline
Ours (with KL)               & 56.44 \textbf{(-6.80)}                           & 53.51 \textbf{(-1.39)}                          & 53.17 \textbf{(-8.96)}                          & 64.55 \textbf{(-6.08)}                          & 52.62 \textbf{(-8.82) }                             & 77.86 \textbf{(+0.90)}     \\
Ours (with 1W) & 53.60 \textbf{(-9.64)} & 53.20 \textbf{(-1.70)} & 57.80 \textbf{(-4.33)} & 69.70 \textbf{(-0.94)} & 60.50 \textbf{(-0.94)} & 75.50 \textbf{(-1.46)}\\
\hline
\end{tabular}}
\end{table*}

\subsection{Sensitivity analysis}
In Table~\ref{table:sensitivity}, we compare the mAP scores obtained for the cross-sensor domain shift, using different combinations of $\alpha_1 $ and $\alpha_2 $  defined in Eq.~\eqref{eq:loss-adv}. We can observe from these results that giving either the same weights to each component or a slightly larger weight to the first component (negative labels) results in better performance.
\begin{table*}[!t]
\caption{mAP comparison of the proposed EM-based GMM clustering with k-means clustering.}
\label{table:kmeans}
\centering
\resizebox{0.7\textwidth}{!}{%
\begin{tabular}{|l|c|c|c|c|c|c|}
\hline
\textbf{Methods} & \textbf{AID$\rightarrow$UCM} & \textbf{UCM$\rightarrow$AID} & \textbf{AID$\rightarrow$DFC} & \textbf{UCM$\rightarrow$DFC} & \textbf{VOC$\rightarrow$Clipart} & \textbf{Clipart$\rightarrow$VOC} \\
\hline
\textbf{Ours}      & 63.24                           & 54.90                           & 62.13                           & 70.64                           & 61.44                & 76.96                \\
\hline
Ours (with k-means)        & 53.58 \textbf{(-9.65)}                           & 52.20 \textbf{(-2.70)}                          & 58.46 \textbf{(-3.68)}                           & 68.06 \textbf{(-2.57) }                         & 49.24 \textbf{(-12.20) }              & 68.27 \textbf{(-8.69)}              \\
\hline
\end{tabular}}
\end{table*}

\begin{table}[!t]
\caption{Sensitivity analysis: A comparison of mAP by varying the values of regularizers for each GMM component on the aerial image datasets.}
\label{table:sensitivity}
\centering
\resizebox{0.45\textwidth}{!}{%
\begin{tabular}{|c|c|c|c|c|}
\hline
\textbf{$\alpha$ values ($\alpha_1, \alpha_2$)} & \textbf{AID$\rightarrow$UCM} & \textbf{UCM$\rightarrow$AID} & \textbf{AID$\rightarrow$DFC} & \textbf{UCM$\rightarrow$DFC} \\
\hline
$\alpha_1$=0.1, $\alpha_2$=0.9                      & 56.0                            & 50.6                            & 55.4                            & 67.1                            \\
$\alpha_1$=0.2, $\alpha_2$=0.8                      & 55.0                            & 52.4                            & 55.3                            & 67.4                            \\
$\alpha_1$=0.3, $\alpha_2$=0.7                      & 56.0                            & 50.3                            & 57.0                            & 69.3                            \\
$\alpha_1$=0.4, $\alpha_2$=0.6                      & 58.0                            & 52.4                            & 59.2                            & 69.3                            \\
$\alpha_1$=0.5, $\alpha_2$=0.5                      & \textbf{63.0}                   & 53.0                            & 57.7                            & \textbf{70.6}                   \\
$\alpha_1$=0.6, $\alpha_2$=0.4                      & 54.4                            & \textbf{54.4}                   & 55.4                            & 66.4                            \\
$\alpha_1$=0.7, $\alpha_2$=0.3                      & 54.9                            & 52.6                            & 57.6                            & 65.6                            \\
$\alpha_1$=0.8, $\alpha_2$=0.2                      & 53.8                            & 52.9                            & \textbf{62.1}                   & 69.3                            \\
$\alpha_1$=0.9, $\alpha_2$=0.1                      & 55.6                            & 53.9                            & 56.0                            & 66.6                            \\
$\alpha_1$=1.0, $\alpha_2$=0.0                      & 57.8                            & 52.0                            & 57.2                            & 69.5       \\
\hline
\end{tabular}}
\end{table}

\subsection{GMM versus k-means}
We employ the popular non-probabilistic clustering technique known as k-means to compare with the used GMM-based clustering. In contrast to the former, which uses hard thresholding to assign data points to specific clusters, GMM uses soft thresholding by maximizing the likelihood that any given data point will be in a given cluster.  Table~\ref{table:kmeans}  compares the mAP scores obtained using the two methods. It can be clearly seen that using k-means results in a significant performance drop for all benchmarks.

\subsection{Distance and divergence measure analysis}
\label{sec:distance}
We propose using the 2-Wasserstein (denoted as 2W) distance as a discrepancy measurement for the learned GMM source and target components. As shown in Table~\ref{table:kl}, we compute the mAP scores using the popular KL-divergence and 1-Wasserstein (1W) distance in lieu of 2W. The effectiveness of the 2W distance as compared to other measures in the proposed method, given its continuity and differentiability properties,  is clearly visible in Table~\ref{table:kl}. More specifically, using the KL divergence or the 1-Wasserstein distance as a discrepancy measure results in a slight to significant reduction in mAP across all benchmarks, ranging from 0.9\% to 9.6\%.

\section{Conclusion}
\label{sec:conclusion}
    In this paper, a discriminator-free UDA approach for MLIC has been proposed. In contrast to existing methods which use an additional discriminator that is trained adversarially, our method leverages the task-specific classifier for implicitly discriminating between source and target domains. This strategy is proposed to avoid decoupling the classification and the discrimination tasks, while reducing the number of required network parameters. To achieve this, the adversarial loss has been redefined using a F\'echet distance between the corresponding GMM components estimated from the classifier probability predictions. We have demonstrated that the proposed approach achieves state-of-the-art results on seven datasets covering three possible areas of domain shift, while considerably decreasing the size of the network. In future works, we will investigate a differentiable strategy for fitting the GMM for a fully end-to-end training of the network.

{\small
\bibliographystyle{ieee_fullname}
\bibliography{egbib}
}

\end{document}